\documentclass[preprint,12pt]{elsarticle}

\pdfoutput=1

\usepackage{graphicx}
\usepackage{amssymb}

\usepackage{lineno}
\usepackage{hyperref}

\usepackage{epsfig}
\usepackage{subfigure}
\usepackage{calc}
\usepackage{amssymb}
\usepackage{amstext}
\usepackage{amsmath}
\usepackage{amsthm}
\usepackage{multicol}
\usepackage{pslatex}

\usepackage{multicol}
\usepackage[english]{babel}
\usepackage[colorinlistoftodos]{todonotes}
\usepackage{algorithm}
\usepackage{algorithmic}
\usepackage{here}
\usepackage{longtable}
\usepackage{a4wide}
\usepackage{graphics}
\usepackage{amsfonts}
\usepackage{array}
\usepackage{url}
\usepackage{subfigure}
\usepackage{here}
\usepackage{tabularx}
\usepackage{booktabs}
\usepackage[flushleft]{threeparttable}

\theoremstyle{definition}

\journal{arXiv}

\begin{document}

\begin{frontmatter}


\title{MIRA: A Computational Neuro-Based Cognitive Architecture Applied to Movie Recommender Systems}



\author{M. B. Santos} 
\author{A. M. Lima}
\author{L. A. Silva}
\author{F. S. Vargas}
\author{G. A. Wachs-Lopes}
\author{P. S. Rodrigues}

\address{Computer Science Department, Centro Universit{\'a}rio FEI, S{\~a}o Paulo, Brazil \\ \{unifmsantos, unifalima, uniflsilva, uniffvargas, gwachs, psergio\}@fei.edu.br}



%

\begin{abstract} 
The human mind is still an unknown process of neuroscience in many aspects. Nevertheless, for decades the scientific community has proposed computational models that try to simulate their parts, specific applications, or their behavior in different situations. The most complete model in this line is undoubtedly the LIDA model, proposed by Stan Franklin \cite{franklin2016lida} with the aim of serving as a generic computational architecture for several applications. The present project is inspired by the LIDA model to apply it to the process of movie recommendation, the model called MIRA (Movie Intelligent Recommender Agent) presented percentages of precision similar to a traditional model when submitted to the same assay conditions. Moreover, the proposed model reinforced the precision indexes when submitted to tests with volunteers, proving once again its performance as a cognitive model, when executed with small data volumes. Considering that the proposed model achieved a similar behavior to the traditional models under conditions expected to be similar for natural systems, it can be said that MIRA reinforces the applicability of LIDA as a path to be followed for the study and generation of computational agents inspired by neural behaviors.
\end{abstract}


\end{frontmatter}

\section{Introduction}
\label{sec:introduction}

Recommender Systems (RS) are intelligent computational architectures which intermediate user experience with database systems offering considerable types of resources through building and giving suggestions to make the user service an efficient and practical involvement. Traditionally, these systems use mechanisms base on Artificial Intelligence and Machine Learning methods in order to process information and present to the user items that may interest them. An efficient RS can generate positive results for some commercial or non-commercial organizations, such as user increasing, sales and the improvement of the user service, where they present the contents that suit the user profile.

The techniques and mathematical-computer models to support these ideas, such as data clustering, Bayesian Statistics and Neural Networks, have been proposed for decades and as the hardware technologies expand these models have been being applied to different challenges, making RS more efficient and known. A common example of tool used to support RS is convolutional networks which although they have been proposed less than a decade, they are based on well known and defined principles for over fifty years.

Perhaps the most well known applications of a RS is for the area of movie recommendation, e-commerce and medical systems. Specifically, this last one for both patients, health professionals and hospitals. In the particular case of movie recommendation, a more popular system is probably the Netflix, which currently supports more than 118 million users worldwide, but yet still having many challenges to overcome.

On the other hand, areas such as neuroscience, psychology, philosophy from one side and artificial intelligence and computer science by another began to develop research on cognitive science, by decades. Inspired by the doctrine of behaviorism in the early twentieth century, philosophy and psychology sought to understand through research the functioning of the mind and its related processes.  At the same time, neuroscience developed research focused uniquely on biological processes. Thus, influenced by both areas, Artificial Intelligence arose with the purpose of representing the mind processes through logical-computational rules.

However, under the new discoveries about human brain behavior, well known challenge faced by Artificial Intelligence  now has being studied through the so-called Cognitive Computational Models.  Generally speaking, they are models based on theories about consciousness, capable of solving problems with perceptions from the external environment as input, also understood and influenced by a range of other perceptions already known to the model.

Among all the developed models the most complete and used is the LIDA (Learning Intelligent Distributed Agent), which is composed by a parallel sequence of modules that form an architecture designed to structure high-level computational problems similar to the behavioral structure of human mind, based only on what is known in this regard so far.

The LIDA ({\it Learning Intelligent Distributed Agent}) is a theoretical cognitive computational model that originated the framework of the same name. This framework has implemented a large part of the proposed modules in the theoretical model and, therefore, among all similar architectures, as already mentioned, it is the most complete that intend to simulate the human mind behavior, being successfully applied in considerable areas of knowledge - as in robotics and medical analysis.

Thus, the contribution of this work is a proposal of modeling a movie recommender system structured as a cognitive model on LIDA architecture \footnote{The source code is avaliable at: \url{https://github.com/VisaoComputacional/mira}}.

The remainder of this paper is organized as following. Section 2 presents a bibliographic review. Section 3 presents the concepts of consciousness. Section 4 explains what cognitive science is. Section 5 comments about Global Workspace Theory. Section 7 presents the LIDA model. Section 8 defines recommendation architectures. Section 9 presents the proposed model. Section 10 shows the results and discussion and Section 11 presents the conclusions.

\section{Related Works}

In recent decades, numerous papers have discussed the concepts of consciousness, attention and cognition \cite{dennett1992time}, \cite{lamme2004separate}, \cite{velmans2009define}, while several others have discussed and suggested that ideas born in neuroscience and psychology can be modeled on machines. 
\cite{perlovsky2006toward} and \cite{tononi2008neural} address these ideas and \cite{mcdermott2007artificial} analyzes how the concept of Artificial Intelligence can be applied to computers. \cite{maclennan2008consciousness} discusses application of consciousness in robots and \cite{manzotti2008artificial} shows some technological and theoretical obstacles to be faced in the field of artificial consciousness. \cite{seth2009explanatory} discusses how consciousness-related theories can best be understood through computational models, and \cite{prasad2010perspective} discusses the role of information in making machines aware, and \cite{wallach2010conceptual} discuss how the LIDA model can be adapted to model rational aspects of cognitive decision making. The authors clains that the LIDA model presented promising results on the subject, demonstrating its efficiency. In \cite{tononi2011integrated} presents ideas to represent the brain both natural and artificial. And \cite{graziano2014mechanistic} approach the attention process, suggesting that this phenomenon can be projected on machines. Also \cite{reggia2016computational} studies consciousness and present possible neuro-computational correlates of consciousness that have been proposed or recognized in recent years based on cognitive architectures of biological inspiration. They also suggest that the study of computational correlates of consciousness will lead to a better understanding of this biological phenomenon.

Parallel to the neuroscience and psychology discoveries, various theoretical models emerged. \cite{mathis1996conscious} propose a computational theory of consciousness to models visual perception. \cite{newman1997neural} presents an awareness model based on known architectures that implanted the GWT (Global Workspace Theory) partially in neural models, in which it presents an understanding of the GWT and its relation with neurology, having as a disadvantage its complexity. Nonetheless, there have been good results in the progress of the GWT studies with neurology. \cite{cleeremans2005computational} points to advanced computational models as a pathway for future research in understanding consciousness. \cite{baars2007architectural} describe a hybrid cognitive architecture based on neural networks aimed at achieving characteristics analogous to some brain functions. The article introduces the concept of Global Workspace Theory (GWT), as well as \cite{baars2009consciousness}, for practical computational predictions. The LIDA algorithm has a GWT-based implementation, demonstrating this in human activities. \cite{graziano2011awareness} argues that the proposed consciousness model is conceptually simple and testable, and \cite{kinouchi2014model} proposes a top-down model involving associative temporal memory, where both consciousness and attention are independent functions. \cite{djerroud2016towards} propose a model of consciousness that represents a multi-agent system capable of evolving, generating data and actions in the environment.

Considering the discoveries in neuroscience and psychology, in addition to the theoretical models proposed, some computational models have arisen based on these theories. \cite{sun1997learning} presents the cognitive model CLARION (Connectionist Learning with Adaptive Rule Induction Online) also tested by \cite{sun1999accounting}, while \cite{franklin1998ida} and \cite{franklin2014lida} describe the architecture of the Intelligent Distribution Agent, the so-called IDA. Other articles that propose new models are \cite{franklin2000modeling}, \cite{johnson2000artificial}, \cite{franklin2003conscious}, which later gave rise to LIDA \cite{franklin2007lida}, \cite{franklin2016lida}, which is presented as a functional model of a conscious machine in its theoretical foundation. The authors describe LIDA as a GWT-based model, its architecture, its memory modules, and its work cycles. In the end, the authors claim that the model presents relevant characteristics of consciousness, but it needs to be improved in order to obtain more cognitive characteristics. Until the publication of this article, the LIDA model had not been applied in any specific area. Also, implementations of cognitive models are proposed in \cite{moreno2007modeling}, \cite{taylor2007codam}, \cite{moreno2008applying}, \cite{starzyk2011computational}, \cite{vassev2013implementing}, \cite{haladjian2016artificial} and \cite{shi2017machine}.

Finally, articles reviewing the area of computational models for cognition can be read in \cite{kozma2007computational}, \cite{sun2007computational}, \cite{gamez2008progress}, \cite{gok2012philosophical} and \cite{reggia2013rise}, where one can read a better evolution from the point view of neural, computational and applications.

\section{Consciousness} 
Consciousness is a intrinsically hard subject. It does not have a certain definition of what is about, but over the years, different definitions were showed, contributing to deeply studies that exceeded philosophy and are applied in manifold areas of science. The Consciousness has many different means, for this, a lot of studies focus in portions of consciousness to resolutions of problems. For example, \citet{baars2002conscious} with The Global Workspace Theory (GWT), proposes that consciousness is the brain gateway. The GWT suggests that consciousness allows different networks to cooperate in solving problems, like specifies items from immediately memory.

\subsection{Functional Consciousness} 
Functional consciousness has an important role as a perceptual filter, which allows the agent just concentrated in more relevant information. Thus, help in action selection, allowing agent gather resources, in order to choose what to do in sequence and solve problems in an efficient way \cite{franklin2003conscious}.

\subsection{Phenomenal Consciousness } 
Phenomenal Consciousness refers to subjective sensation of conscious experience \cite{sun1997learning}. The knowledge of color or sound is an inexplicable experience, for example, it is not possible to explain to a person who has no vision from birth what is the purple color, for phenomenal consciousness is impossible to be transmitted in words but must be experienced to be understood \cite{franklin2003conscious}.

\subsection{Artificial Consciousness} 
Machine consciousness or artificial consciousness is inspired in philosophy, psychology and neuroscience, as well as share many of the goals of artificial intelligence.  Thus, is not a unified field with set of objectives clearly defined. Currently, researchers in different areas work in distinct aspects of problem, what difficult oftentimes comprehension of how everything works \cite{gamez2008progress}. One of many possible definitions can be found in \cite{prasad2010perspective}: "A machine is conscious if beyond necessary mechanisms to perception, action, learning and associative memory, it has an executive central which controls all machine processes (conscious and unconscious).

\section{Cognitive Science} 
Cognitive Science \cite{sun1994computational} is a study object of different areas, between them computer science. Cognition is the information processing through senses to obtain knowledge, and the purpose of the study of cognitive science is to understand the functioning of the human mind.

The main type of approach of this article is symbolic, which consists in the cognition being able to be represented by formulas and mathematical modulations, thus, can be replicated in computational models.

\section{Attention} 
Attention is a strongly concept related to consciousness, \cite{graziano2014speculations} speculates that consciousness has a biological arisement as from attention, being this concept used every day frequently as mental concentration about something specific. However, to neuroscience \cite{graziano2014speculations,graziano2015attention} attention is the information selection cognitive process. The informations are processed by a competition to select which of these will be deeply performed, being the competition known as "biased competition". The involved signals in this competition can be influenced by \textit{top-down} and \textit{bottom-up} stimuli, once the stimulus be attended this one wins the competition. Nevertheless, the attention is a state in constant change that occurs sometimes of a visual representation win the competition of moment.

\section{Global Workspace Theory} 
\label{gw-theory}
The Global Workspace Theory (GWT) is a cognitive architecture model, proposed by Baars in \cite{baars1997theatre}, with the goal to explain conscious and unconscious process that occurs in brain. The theory suggests that consciousness can be related to ''limited capability of brain"; in other words, immediately memory and selective capability of attention, since this limited capability comes from a slight impression where the brain is a slow organ which executes tasks sequentially. However, when it is actually studied, in fact, that the brain is an organ with a neural networks complex set, working in parallel with each network having its own cognitive specialty. GWT considers consciousness as global access or the act of transmission between cognitive process and memories. It can be compared  with an old artificial intelligence concept on which it was based, the "blackboard". This concept is a common basis for different specific processes.

\begin{figure}[!htb]
  \centering 
  \includegraphics[height=0.35\textheight, keepaspectratio]{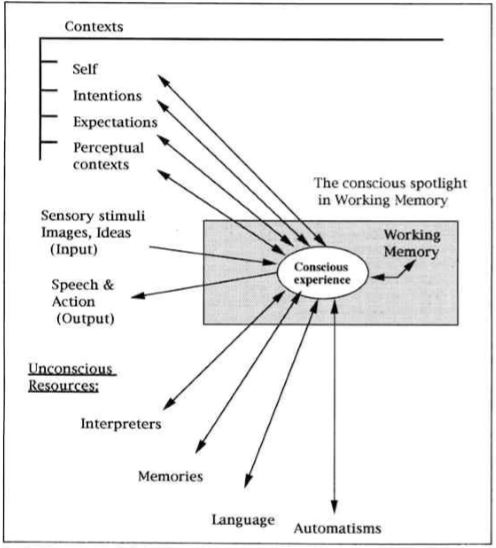}
  \caption{\textit{Global Workspace} diagram}
  \caption{Source: \cite{baars1997theatre,baars2005global}.}
  \label{fig:Global_Workspace}
\end{figure}

The functioning of this theory can be exemplified by the theater metaphor of consciousness or mind, seen in Figure \ref{fig:Global_Workspace}, in which the theater is dark and unconscious. Consciousness is taken as bright spots on stage, guided by a spotlight of attention in an executive leadership. Attention spotlight is the guide to consciousness, both voluntarily and spontaneously. Therefore, only events that are in this luminous area are strictly conscious, and there is a competition between the thoughts, sensations or images, in order to reach the stage and the luminous part. Such that, the winner of this competition becomes conscious. In backstage of theater are contexts that are often unconscious and strongly influence conscious processes, shaping the conscious events. The director works invisibly behind the scenes, being the one who makes the decisions based on goal. Finally, we have the public, in which consciousness is a gateway to a vast area of unconscious cognitive processes, such as long-term memory, language, interpreters and automatisms.

\section{LIDA Architecture}
\label{LIDA}
The Learning IDA (LIDA) is a computational cognitive model presented by Stan Franklin in \cite{franklin2016lida}, taken demonstrated as an evolution of the IDA model, where three types of learning were added: perceptual, episodic and procedural. The LIDA cognitive model is conceptual and partly computational, as well it is divided into modules that has the purpose of perceiving the environment and acting according to its own schedule.
These modules working together form a cognitive atom, where different high-level cognitive processes are performed. However, in order to be able to act in an environment, many cognitive processes are necessary and, for this, the cognitive atoms are processed continuously in the LIDA's Cognitive Cycle.

\begin{figure}[!htb]
	\centering
	\includegraphics[height=0.15\textheight, keepaspectratio]{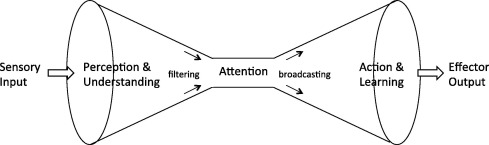}
	\caption{Schematic diagram of the cognitive cycle of the LIDA model}
	\caption{Source: \cite{franklin2016lida}.}
	\label{fig:Ciclo_Cognitivo}
\end{figure}

The LIDA Cognitive Cycle is a cycle that allows frequent sampling and responses. For this to be possible it is divided into three phases: the perception and understanding phase, the attention phase and the action phase (Figure \ref{fig:Ciclo_Cognitivo}). The first phase uses external data to understand the current situation, the second phase filters content according to its degree of relevance and propagates globally, and the third phase selects an appropriate response, executes that response and learns into a memory systems. Each of these phases is divided into processes that work in parallel, with the exception of the process of awareness and action selection.

\begin{figure}[!htb]
  \centering
  \includegraphics[width=16cm]
  {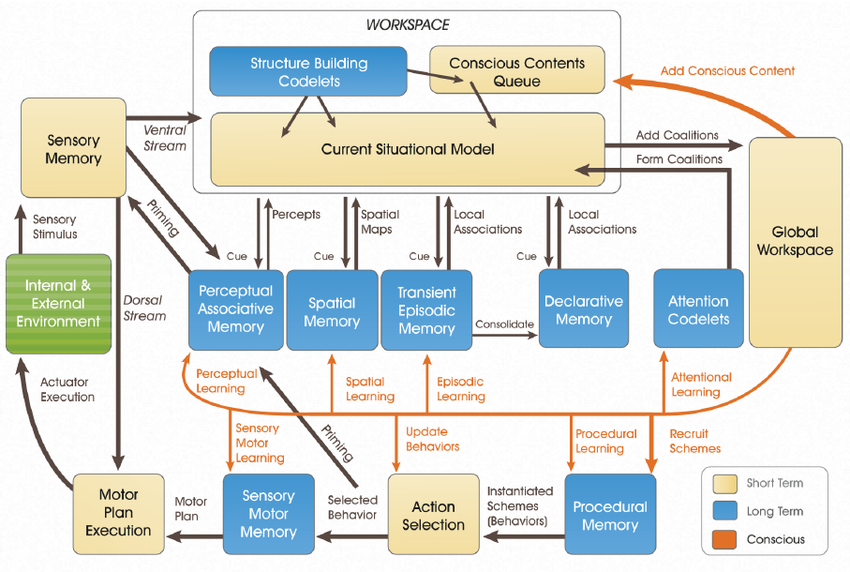}
  \caption{LIDA's cognitive cycle}
  \caption{Source: \cite{franklin2016lida}.}
  \label{fig:Ciclo_Cognitivo_Modular}
\end{figure}

The Perception and Understanding phase starts from external and internal stimuli sent to Sensory Memory (Figure \ref{fig:Ciclo_Cognitivo_Modular}). The resulting content generates perception by passing through the Perceptual Associative Memory (Figure \ref{fig:Ciclo_Cognitivo_Modular}) and is made available in the Current Situational Model (Figure \ref{fig:Ciclo_Cognitivo_Modular}). The Current Situational Model (Figure \ref{fig:Ciclo_Cognitivo_Modular}) is updated by feeding Perceptual Associative Memory (Figure \ref{fig:Ciclo_Cognitivo_Modular}), Spatial Memory (Figure \ref{fig:Ciclo_Cognitivo_Modular}), Transient Episodic Memory (Figure \ref{fig:Ciclo_Cognitivo_Modular}), and Declarative Memory (Figure \ref{fig:Ciclo_Cognitivo_Modular}). Likewise, updates in Workspace (Figure \ref{fig:Ciclo_Cognitivo_Modular}) are done by Structure Building Contents (Figure \ref{fig:Ciclo_Cognitivo_Modular}) using data from the Current Situational Model (Figure \ref{fig:Ciclo_Cognitivo_Modular}) and Conscious Contents Queue (Figure \ref{fig:Ciclo_Cognitivo_Modular}), generating what is called preconscious state in the agent.

During the attention phase, the Attention Codelets (Figure \ref{fig:Ciclo_Cognitivo_Modular}) evaluate the Current Situational Model (Figure \ref{fig:Ciclo_Cognitivo_Modular}) in search of selecting content to be brought to consciousness. When it find such content, an alliance is created to compete for access to consciousness and the winning alliance will have its content propagated globally.

In the action and learning phase most of the LIDA modules select contents of the process of consciousness that are appropriate to their learning, select behaviors stored in Procedural Memory (Figure \ref{fig:Ciclo_Cognitivo_Modular}) that accurately respond to input stimuli and then the module Action Selection (Figure \ref{fig:Ciclo_Cognitivo_Modular}) will select a behavior to be sent to the Sensory Motor Plan (Figure \ref{fig:Ciclo_Cognitivo_Modular}). It creates or selects a suitable motor plane to execute. Finishing the LIDA cognitive cycle.

\section{Recommender Systems} 
\label{sistema-recomendacao-trad}

The recommender systems field emerged in 1990s and since then evolved as the computerization of sectors such as e-commerce, social networks and multimedia become more present in daily lives of users of these services.

Commercial and scientific developments of recommender systems have brought advantages for the growth of organizations, data analysis, and decision making, resulting in appropriate suggestions of items to their users' profile, online sales growth, content organization, among other advances \cite{isinkaye2015recommendation}.

The main techniques described in \cite{burke2002hybrid} research are:
\begin{enumerate}
    \item \textbf{Content-based Filtering:} the suggested recommendations are according to items that were of interest to the user in the past.
    \item \textbf{Collaborative Filtering:} relates suggestions for a given user based on users that contain similar profiles.
    \item \textbf{Community-based:} uses as reference to make proposals the profile of friends and people linked to the user.
    \item \textbf{Hybrid recommender systems:} consists of merging two or more recommendation techniques.
\end{enumerate}

\section{Proposed Methodology} 
\label{metodologia}
\begin{figure}[!htb]
  \centering 
  \includegraphics[width=16cm]{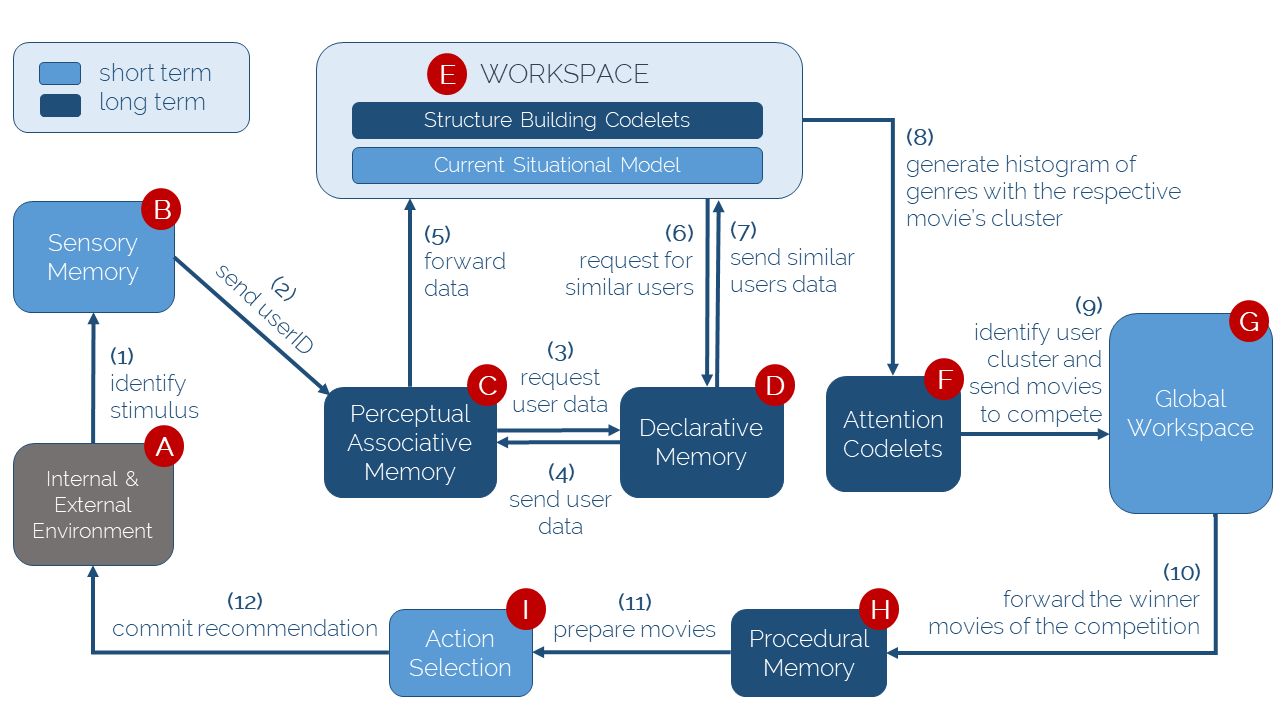}
  \caption{MIRA Model}
  \label{fig:MIRA_Model}
\end{figure}

The proposed methodology for a recommender movie system agent implementation is tightly inspired in LIDA model, so-called MIRA (Movie Intelligent Recommender Agent). The MIRA model is composed of nine modules, with short and long term memories, as Figure \ref{fig:MIRA_Model} illustrates.

In general, the cognitive cycle of the proposed model of this work, recognizes input data (Bloc A and B) in Figure \ref{fig:MIRA_Model}, process movie data (Bloc C, D and E) in Figure \ref{fig:MIRA_Model}, mounts the processed information based on attention concepts (Bloc F and G) in Figure \ref{fig:MIRA_Model} to be resend to the external environment (Bloc H and I) in Figure \ref{fig:MIRA_Model}.

The MIRA cognitive model is based of the following twelve linear steps.

\subsection{Stimulus Identification}
Firstly, the Internal and External Environment constantly observes the users actions in the system. These information is sent to Sensory Memory in stimulus means. In the MIRA modeling, the stimulus is the user identification, an integer number called \textit{userID}.

\subsection{UserID Sending}
The Sensory Memory modules captures the user number identification and send it to the Perceptual Associative Memory (PAM) module.

\subsection{User Data Request}
The PAM module requests for the \textit{userID} data to Declarative Memory, which stores all the movies related to the users in its database.

\subsection{User Data Sending}
The Declarative Memory searchs for the \textit{userID} data and reply the PAM request with the user content.

\subsection{User Data Routing}
The PAM receives the answer of Declarative Memory module in dataframe structure and forwards to Workspace, which masters the building of the structures before the attention methods.

\subsection{Similar Users Request}
When the user data comes to Workspace, the module makes a request for Declarative Memory to search for $N$ similar users in relation to \textit{userID}.

\subsection{Similar Users Sending}
The Declarative Memory applies the cosine similarity technique to find similar users. With this information, the module also formats all similar users data - as watched movies, ratings and genres - and sends back to Workspace.

\subsection{Histogram Generation}
The users information sent by Declarative Memory are used by Workspace, which generates a histogram of all movies listed from similar users watched movies. The histogram is generated splitting every kind of genre from genres list and assign to a binary value, where 1 indicates that the genre is present, and 0 that the genre does not exist in the movie. With these values combination, the K-Means algorithm  is used to identify which cluster every movie belongs. 

\subsection{Cluster Identification}
In this step, the Attention Codelets receives user historical data and recognize which movie cluster the user more likely to watch, with the application of K-Means technique. With the histogram generated in previously step, the Attention Codelets captures only the movies with the same cluster of the \textit{userID}. 

\subsection{Movie Competition}
The competition phase starts when Attention Codelets sends to Global Workspace module the movies with the user cluster based on similar users data. The average rating of every movie is computed, and the movies with the highest values wins the competition.

\subsection{Movie Preparation}
After the competition of movies, the winners are sent to Procedural Memory module, which prepares the movies that are going to be recommended and assign a title to the selected cluster.

\subsection{Recommendation Commit}
The list of movie recommendations is ready and prepared for the user. The final step consists in showing the movies to the user, ranked from the highest rating value to the lower.



\section{Results and Discussion}
\label{Resultados}
The cognitive model MIRA in this section was aimed to several experiments and comparisons with a traditional recommender model.

\subsection{Clusters and similar user numbers}
The first experiment was dedicated to defining the best combination among clusters in K-Means algorithm, used mainly in Workspace to determine best cluster to main user, and similar user numbers, used in declarative memory to be defined how many users will be considers resembling to the main user. Where K is the number of clusters, was generated in experiment: $k$ = 5, $k$ = 6, $k$ = 7, $k$ = 8, $k$ = 9 and $k$ = 10. Just as they were generated 5, 10, 20, 30, 40 and 50 similar users to the main user to run the experiment.

It was selected 10 main users to execute this experiment, for each user being executed a different combination, for example, to user 1 each clusters number defined above to different similar user numbers were tested. The same was applied to other users. For all users 40 movies were recommended, every result was analyzed to define what combination between $K$ and similar user numbers is correctly recommending the movies for a given user.

The results validation was realized considering user rating to certain movies, using movielens 1 MM database. The 6 genres most rated with a bigger or equal to 4 rating, it will be considering user preferred genres. For this reason, the 40 movies recommended by the system, for every combination between K and similar user numbers, are considered correct if they belong to one of these 6 genres, else are considered incorrect. The precision index is then calculated using the total number of films that are correctly recommended by the number of recommended films.

It was observed in this experiment that MIRA works better when $K$ number is equal to 8, regardless of similar user numbers, because it presents higher precision index, when we look at similar user numbers that best recommend with $K$ equal to 8, we conclude that comparing 10 similar users to the main user, MIRA stay more precise. Table \ref{tab:tab_cluster_7_8_grupo} shows the precision numbers.

\begin{table}[H]
\centering
\begin{tabular}{|c|c|c|}
\hline
Cluster & 7    & 8    \\ \hline
10      & 24\% & 26\% \\ \hline
20      & 19\% & 23\% \\ \hline
30      & 22\% & 23\% \\ \hline
\end{tabular}
\caption{Table with the mean precision percentages of clusters 7 and 8 for 10, 20 and 30 similar users}
\label{tab:tab_cluster_7_8_grupo}
\end{table}

\subsection{MIRA vs TRADITIONAL}
\label{MIRA vs Tradicional}
Aiming to generate an intelligent agent based on LIDA model applied in recommender systems of movies, this experiment intends to demonstrate that proposed model MIRA meets similar expectations of a traditional models. So, it was used a model implemented in \cite{TCCMiller,2018-zamberlan} which consists in a predictive movie assessment system.

MIRA model was executed with $K$ equal to 8, where $K$ is the number of clusters, and 10 similar users to main user like previously established, for presents best perform as to precision in your results.

\begin{figure}[htbp]
\centering
\begin{minipage}[b]{8cm}
    \centering
    \includegraphics[width=1.0\linewidth, height=1.0\linewidth]{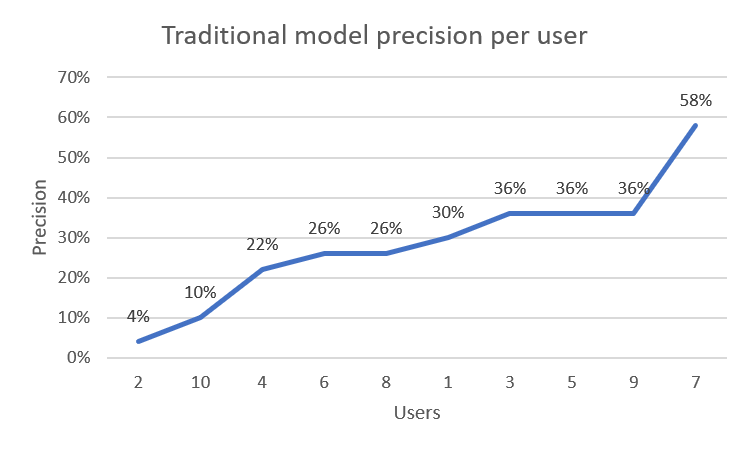}
    (a)
\end{minipage}
\begin{minipage}[b]{8cm}
    \centering
    \includegraphics[width=1.0\linewidth, height=1.0\linewidth]{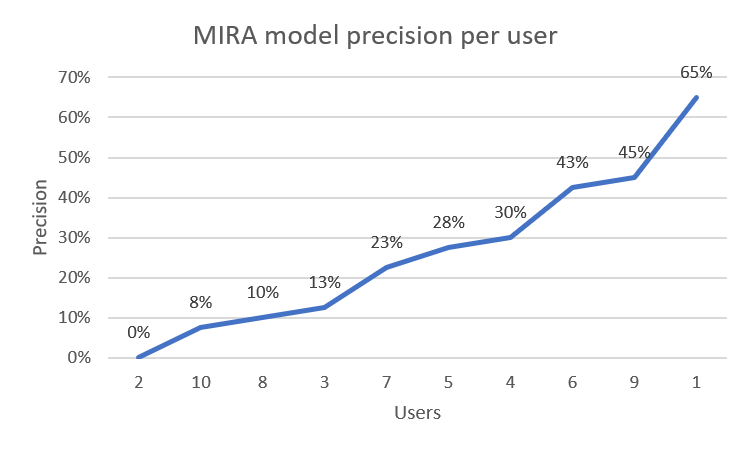}
    (b)
\end{minipage}
\caption{(a): Precision analyze graph of traditional model per user with precision average = 28\%. (b): Precision analyze graph of MIRA model per user with precision average = 26\%} 
\label{fig:Grafico_Precisao}\end{figure}

The comparison metric between models was the same of section before, which was replicated on traditional model. Thus, it can be seen in the graphs that presents the precision results for each model in Figures \ref{fig:Grafico_Precisao}-(a) and \ref{fig:Grafico_Precisao}-(b), their precision behaviour compared to the 10 users analyzed. It is possible to observe average precision percentage of MIRA model with 26\% (Figure \ref{fig:Grafico_Precisao}-(b)), which is very close to the percentage presented by traditional model with a 28\% average precision (Figure \ref{fig:Grafico_Precisao}-(a)). Among these results, the MIRA model establishes what may be a possible solution to the traditional models of recommendation.

\subsection{MIRA evaluated by users}
In order to reinforce that proposed cognitive model in this article are in fact a viable solution to movies recommendation, the experiment below uses movie ratings of three volunteers to the movies they already watched and after defines satisfaction rate regarding to proposed recommendations by MIRA model.

To carry out the experiment, it was requested to each volunteer to rate every movie ever then watched them, in a database with 1642 movies. Then, the records of these evaluations were inserted into the model database, to generate recommendations for each volunteer. From the recommendations obtained for each user, hit rates are shown in Table \ref{tab:tab_teste2_pessoal_perc}.

It was possible to observe that in this test, the average precision (31,4\%) is greater than the precision of MIRA model in Section test \ref{MIRA vs Tradicional}, which represents a precision gain of 6,4\%. That is, when the user evaluates the movies recommended by MIRA, he is much more satisfied with the recommendation than when we only analyze what has already been evaluated by him. The cluster selected by MIRA model was appropriate according to the users who participated in this experimental phase; that is, the categories of movies were of the profile of the users.

\begin{table}[htb]
\centering
\begin{tabular}{|c|c|}
\hline
Experiment          & Precision (\%) \\ \hline
Volunteer (A)   & 30 \%       \\ \hline
Volunteer (B)   & 48 \%       \\ \hline
Volunteer (C)   & 20 \%       \\ \hline
\textbf{Average} & 32,7 \%     \\ \hline
\end{tabular}
\caption{Precision among users who evaluated MIRA's recommendations based on their watched movies}
\label{tab:tab_teste2_pessoal_perc}
\end{table}

\section{Conclusions}
The proposed work in this article had as goal to generate a movie recommendation agent inspired by cognitive architecture of LIDA, describe in Section \ref{LIDA}. Based on cognitive modeling of LIDA, MIRA was adapted attributing tasks to generate movie recommendation, according to long and short-term memory functionalities.

For this purpose, the MIRA model was separately implemented for each memory, that was presented in Figure \ref{fig:MIRA_Model}, using techniques describes in Section \ref{metodologia}, developing information flow between modules sequentially.

Thus, the approach of a recommender system modeling through a cognitive architecture can be considered practicable to this application to represents balanced results according to established metrics and results obtained in Section \ref{Resultados}.

However, it can be stated that the proposed cognitive model MIRA, being the first cognitive model proposed in literature to movie recommendation task, is closer to a real cognitive behavior than the traditional model. Thus, assuming its great architecture variabilities, it is possible in the future, when a neural system behavior would be reported, artificial cognitive models like MIRA, can become great alternatives both to better understand neural models and to simulate these models.

\section*{Acknowledgment} The authors are thank to FAPESP (Foundation for Research of the S\~{a}o Paulo State, Brazil), CNPq (National Council for Scientific and Technological Development, Brazil), CAPES (Coordination of High Level Graduation, Brazil) and FEI (Foundation for Ignatian Education, S\~{a}o Paulo, Brazil) for any direct or indirect support they have given to this research project.

\bibliographystyle{model1-num-names}
\bibliography{refs.bib}

\begin{thebibliography}{53}
\expandafter\ifx\csname natexlab\endcsname\relax\def\natexlab#1{#1}\fi
\providecommand{\bibinfo}[2]{#2}
\ifx\xfnm\relax \def\xfnm[#1]{\unskip,\space#1}\fi
\bibitem[{Franklin et~al.(2016)Franklin, Madl, Strain, Faghihi, Dong, Kugele,
  Snaider, Agrawal, and Chen}]{franklin2016lida}
\bibinfo{author}{S.~Franklin}, \bibinfo{author}{T.~Madl},
  \bibinfo{author}{S.~Strain}, \bibinfo{author}{U.~Faghihi},
  \bibinfo{author}{D.~Dong}, \bibinfo{author}{S.~Kugele},
  \bibinfo{author}{J.~Snaider}, \bibinfo{author}{P.~Agrawal},
  \bibinfo{author}{S.~Chen},
\newblock \bibinfo{title}{A lida cognitive model tutorial},
\newblock \bibinfo{journal}{Biologically Inspired Cognitive Architectures}
  \bibinfo{volume}{16} (\bibinfo{year}{2016}) \bibinfo{pages}{105--130}.
\bibitem[{Dennett and Kinsbourne(1992)}]{dennett1992time}
\bibinfo{author}{D.~C. Dennett}, \bibinfo{author}{M.~Kinsbourne},
\newblock \bibinfo{title}{Time and the observer: The where and when of
  consciousness in the brain},
\newblock \bibinfo{journal}{Behavioral and Brain Sciences} \bibinfo{volume}{15}
  (\bibinfo{year}{1992}) \bibinfo{pages}{183--201}.
\bibitem[{Lamme(2004)}]{lamme2004separate}
\bibinfo{author}{V.~A. Lamme},
\newblock \bibinfo{title}{Separate neural definitions of visual consciousness
  and visual attention; a case for phenomenal awareness},
\newblock \bibinfo{journal}{Neural networks} \bibinfo{volume}{17}
  (\bibinfo{year}{2004}) \bibinfo{pages}{861--872}.
\bibitem[{Velmans(2009)}]{velmans2009define}
\bibinfo{author}{M.~Velmans},
\newblock \bibinfo{title}{How to define consciousness: And how not to define
  consciousness},
\newblock \bibinfo{journal}{Journal of Consciousness Studies}
  \bibinfo{volume}{16} (\bibinfo{year}{2009}) \bibinfo{pages}{139--156}.
\bibitem[{Perlovsky(2006)}]{perlovsky2006toward}
\bibinfo{author}{L.~I. Perlovsky},
\newblock \bibinfo{title}{Toward physics of the mind: Concepts, emotions,
  consciousness, and symbols},
\newblock \bibinfo{journal}{Physics of Life Reviews} \bibinfo{volume}{3}
  (\bibinfo{year}{2006}) \bibinfo{pages}{23--55}.
\bibitem[{Tononi and Koch(2008)}]{tononi2008neural}
\bibinfo{author}{G.~Tononi}, \bibinfo{author}{C.~Koch},
\newblock \bibinfo{title}{The neural correlates of consciousness},
\newblock \bibinfo{journal}{Annals of the New York Academy of Sciences}
  \bibinfo{volume}{1124} (\bibinfo{year}{2008}) \bibinfo{pages}{239--261}.
\bibitem[{McDermott(2007)}]{mcdermott2007artificial}
\bibinfo{author}{D.~McDermott},
\newblock \bibinfo{title}{Artificial intelligence and consciousness},
\newblock \bibinfo{journal}{The Cambridge handbook of consciousness}
  (\bibinfo{year}{2007}) \bibinfo{pages}{117--150}.
\bibitem[{MacLennan(2008)}]{maclennan2008consciousness}
\bibinfo{author}{B.~J. MacLennan},
\newblock \bibinfo{title}{Consciousness: Natural and artificial},
\newblock \bibinfo{journal}{Synthesis philosophica} \bibinfo{volume}{22}
  (\bibinfo{year}{2008}) \bibinfo{pages}{401--433}.
\bibitem[{Manzotti and Tagliasco(2008)}]{manzotti2008artificial}
\bibinfo{author}{R.~Manzotti}, \bibinfo{author}{V.~Tagliasco},
\newblock \bibinfo{title}{Artificial consciousness: A discipline between
  technological and theoretical obstacles},
\newblock \bibinfo{journal}{Artificial intelligence in medicine}
  \bibinfo{volume}{44} (\bibinfo{year}{2008}) \bibinfo{pages}{105--117}.
\bibitem[{Seth(2009)}]{seth2009explanatory}
\bibinfo{author}{A.~Seth},
\newblock \bibinfo{title}{Explanatory correlates of consciousness: theoretical
  and computational challenges},
\newblock \bibinfo{journal}{Cognitive Computation} \bibinfo{volume}{1}
  (\bibinfo{year}{2009}) \bibinfo{pages}{50--63}.
\bibitem[{Prasad and Starzyk(2010)}]{prasad2010perspective}
\bibinfo{author}{D.~K. Prasad}, \bibinfo{author}{J.~A. Starzyk},
\newblock \bibinfo{title}{A perspective on machine consciousness},
\newblock in: \bibinfo{booktitle}{Intl. Conf. Advanced Cognitive Technologies
  and Applications}, \bibinfo{organization}{Citeseer}, pp.
  \bibinfo{pages}{109--114}.
\bibitem[{Wallach et~al.(2010)Wallach, Franklin, and
  Allen}]{wallach2010conceptual}
\bibinfo{author}{W.~Wallach}, \bibinfo{author}{S.~Franklin},
  \bibinfo{author}{C.~Allen},
\newblock \bibinfo{title}{A conceptual and computational model of moral
  decision making in human and artificial agents},
\newblock \bibinfo{journal}{Topics in cognitive science} \bibinfo{volume}{2}
  (\bibinfo{year}{2010}) \bibinfo{pages}{454--485}.
\bibitem[{Tononi(2011)}]{tononi2011integrated}
\bibinfo{author}{G.~Tononi},
\newblock \bibinfo{title}{The integrated information theory of consciousness:
  an updated account},
\newblock \bibinfo{journal}{Archives italiennes de biologie}
  \bibinfo{volume}{150} (\bibinfo{year}{2011}) \bibinfo{pages}{56--90}.
\bibitem[{Graziano and Webb(2014)}]{graziano2014mechanistic}
\bibinfo{author}{M.~S. Graziano}, \bibinfo{author}{T.~W. Webb},
\newblock \bibinfo{title}{A mechanistic theory of consciousness},
\newblock \bibinfo{journal}{International Journal of Machine Consciousness}
  \bibinfo{volume}{6} (\bibinfo{year}{2014}) \bibinfo{pages}{163--176}.
\bibitem[{Reggia et~al.(2016)Reggia, Katz, and Huang}]{reggia2016computational}
\bibinfo{author}{J.~A. Reggia}, \bibinfo{author}{G.~Katz},
  \bibinfo{author}{D.-W. Huang},
\newblock \bibinfo{title}{What are the computational correlates of
  consciousness?},
\newblock \bibinfo{journal}{Biologically Inspired Cognitive Architectures}
  \bibinfo{volume}{17} (\bibinfo{year}{2016}) \bibinfo{pages}{101--113}.
\bibitem[{Mathis and Mozer(1996)}]{mathis1996conscious}
\bibinfo{author}{D.~Mathis}, \bibinfo{author}{M.~C. Mozer},
\newblock \bibinfo{title}{Conscious and unconscious perception: A computational
  theory},
\newblock in: \bibinfo{booktitle}{Proceedings of the eighteenth annual
  conference of the Cognitive Science Society}, pp. \bibinfo{pages}{324--328}.
\bibitem[{Newman et~al.(1997)Newman, Baars, and Cho}]{newman1997neural}
\bibinfo{author}{J.~Newman}, \bibinfo{author}{B.~J. Baars},
  \bibinfo{author}{S.-B. Cho},
\newblock \bibinfo{title}{A neural global workspace model for conscious
  attention},
\newblock \bibinfo{journal}{Neural Networks} \bibinfo{volume}{10}
  (\bibinfo{year}{1997}) \bibinfo{pages}{1195--1206}.
\bibitem[{Cleeremans(2005)}]{cleeremans2005computational}
\bibinfo{author}{A.~Cleeremans},
\newblock \bibinfo{title}{Computational correlates of consciousness},
\newblock \bibinfo{journal}{Progress in brain research} \bibinfo{volume}{150}
  (\bibinfo{year}{2005}) \bibinfo{pages}{81--98}.
\bibitem[{Baars and Franklin(2007)}]{baars2007architectural}
\bibinfo{author}{B.~J. Baars}, \bibinfo{author}{S.~Franklin},
\newblock \bibinfo{title}{An architectural model of conscious and unconscious
  brain functions: Global workspace theory and ida},
\newblock \bibinfo{journal}{Neural Networks} \bibinfo{volume}{20}
  (\bibinfo{year}{2007}) \bibinfo{pages}{955--961}.
\bibitem[{Baars and Franklin(2009)}]{baars2009consciousness}
\bibinfo{author}{B.~J. Baars}, \bibinfo{author}{S.~Franklin},
\newblock \bibinfo{title}{Consciousness is computational: The lida model of
  global workspace theory},
\newblock \bibinfo{journal}{International Journal of Machine Consciousness}
  \bibinfo{volume}{1} (\bibinfo{year}{2009}) \bibinfo{pages}{23--32}.
\bibitem[{Graziano and Kastner(2011)}]{graziano2011awareness}
\bibinfo{author}{M.~S. Graziano}, \bibinfo{author}{S.~Kastner},
\newblock \bibinfo{title}{Awareness as a perceptual model of attention},
\newblock \bibinfo{journal}{Cognitive neuroscience} \bibinfo{volume}{2}
  (\bibinfo{year}{2011}) \bibinfo{pages}{125--127}.
\bibitem[{Kinouchi(2014)}]{kinouchi2014model}
\bibinfo{author}{Y.~Kinouchi},
\newblock \bibinfo{title}{A model of consciousness and attention aimed at
  speedy autonomous adaptation},
\newblock \bibinfo{journal}{Procedia Computer Science} \bibinfo{volume}{41}
  (\bibinfo{year}{2014}) \bibinfo{pages}{120--125}.
\bibitem[{Djerroud and Cherif(2016)}]{djerroud2016towards}
\bibinfo{author}{H.~Djerroud}, \bibinfo{author}{A.~A. Cherif},
\newblock \bibinfo{title}{Towards a computational model of consciousness:
  Global abstraction workspace},
\newblock \bibinfo{journal}{Development} \bibinfo{volume}{747}
  (\bibinfo{year}{2016}) \bibinfo{pages}{63918}.
\bibitem[{Sun(1997)}]{sun1997learning}
\bibinfo{author}{R.~Sun},
\newblock \bibinfo{title}{Learning, action and consciousness: A hybrid approach
  toward modelling consciousness},
\newblock \bibinfo{journal}{Neural Networks} \bibinfo{volume}{10}
  (\bibinfo{year}{1997}) \bibinfo{pages}{1317--1331}.
\bibitem[{Sun(1999)}]{sun1999accounting}
\bibinfo{author}{R.~Sun},
\newblock \bibinfo{title}{Accounting for the computational basis of
  consciousness: A connectionist approach},
\newblock \bibinfo{journal}{Consciousness and Cognition} \bibinfo{volume}{8}
  (\bibinfo{year}{1999}) \bibinfo{pages}{529--565}.
\bibitem[{Franklin et~al.(1998)Franklin, Kelemen, and
  McCauley}]{franklin1998ida}
\bibinfo{author}{S.~Franklin}, \bibinfo{author}{A.~Kelemen},
  \bibinfo{author}{L.~McCauley},
\newblock \bibinfo{title}{Ida: A cognitive agent architecture},
\newblock in: \bibinfo{booktitle}{Systems, Man, and Cybernetics, 1998. 1998
  IEEE International Conference on}, volume~\bibinfo{volume}{3},
  \bibinfo{organization}{IEEE}, pp. \bibinfo{pages}{2646--2651}.
\bibitem[{Franklin et~al.(2014)Franklin, Madl, D'Mello, and
  Snaider}]{franklin2014lida}
\bibinfo{author}{S.~Franklin}, \bibinfo{author}{T.~Madl},
  \bibinfo{author}{S.~D'Mello}, \bibinfo{author}{J.~Snaider},
\newblock \bibinfo{title}{Lida: A systems-level architecture for cognition,
  emotion, and learning},
\newblock \bibinfo{journal}{IEEE Transactions on Autonomous Mental Development}
  \bibinfo{volume}{6} (\bibinfo{year}{2014}) \bibinfo{pages}{19--41}.
\bibitem[{Franklin(2000)}]{franklin2000modeling}
\bibinfo{author}{S.~Franklin},
\newblock \bibinfo{title}{Modeling consciousness and cognition in software
  agents},
\newblock in: \bibinfo{booktitle}{Proceedings of the Third International
  Conference on Cognitive Modeling, Groeningen, NL},
  \bibinfo{organization}{Citeseer}.
\bibitem[{Johnson et~al.(2000)Johnson, Caulfield, and
  Taylor}]{johnson2000artificial}
\bibinfo{author}{J.~L. Johnson}, \bibinfo{author}{H.~J. Caulfield},
  \bibinfo{author}{J.~R. Taylor},
\newblock \bibinfo{title}{Artificial consciousness algorithm for an autonomous
  system},
\newblock in: \bibinfo{booktitle}{Neural Networks, 2000. IJCNN 2000,
  Proceedings of the IEEE-INNS-ENNS International Joint Conference on},
  volume~\bibinfo{volume}{5}, \bibinfo{organization}{IEEE}, pp.
  \bibinfo{pages}{635--640}.
\bibitem[{Franklin(2003)}]{franklin2003conscious}
\bibinfo{author}{S.~Franklin},
\newblock \bibinfo{title}{A conscious artifact?},
\newblock \bibinfo{journal}{Journal of Consciousness Studies}
  \bibinfo{volume}{10} (\bibinfo{year}{2003}) \bibinfo{pages}{47--66}.
\bibitem[{Franklin et~al.(2007)Franklin, Ramamurthy, D'Mello, McCauley, Negatu,
  Silva, and Datla}]{franklin2007lida}
\bibinfo{author}{S.~Franklin}, \bibinfo{author}{U.~Ramamurthy},
  \bibinfo{author}{S.~K. D'Mello}, \bibinfo{author}{L.~McCauley},
  \bibinfo{author}{A.~Negatu}, \bibinfo{author}{R.~L. Silva},
  \bibinfo{author}{V.~Datla},
\newblock \bibinfo{title}{Lida: A computational model of global workspace
  theory and developmental learning.}  (\bibinfo{year}{2007}).
\bibitem[{Moreno et~al.(2007)Moreno, Espino, and
  De~Miguel}]{moreno2007modeling}
\bibinfo{author}{R.~A. Moreno}, \bibinfo{author}{A.~L. Espino},
  \bibinfo{author}{A.~S. De~Miguel},
\newblock \bibinfo{title}{Modeling consciousness for autonomous robot
  exploration},
\newblock in: \bibinfo{booktitle}{International Work-Conference on the
  Interplay Between Natural and Artificial Computation},
  \bibinfo{organization}{Springer}, pp. \bibinfo{pages}{51--60}.
\bibitem[{Taylor(2007)}]{taylor2007codam}
\bibinfo{author}{J.~G. Taylor},
\newblock \bibinfo{title}{Codam: A neural network model of consciousness},
\newblock \bibinfo{journal}{Neural Networks} \bibinfo{volume}{20}
  (\bibinfo{year}{2007}) \bibinfo{pages}{983--992}.
\bibitem[{Moreno and de~Miguel(2008)}]{moreno2008applying}
\bibinfo{author}{R.~A. Moreno}, \bibinfo{author}{A.~S. de~Miguel},
\newblock \bibinfo{title}{Applying machine consciousness models in autonomous
  situated agents},
\newblock \bibinfo{journal}{Pattern Recognition Letters} \bibinfo{volume}{29}
  (\bibinfo{year}{2008}) \bibinfo{pages}{1033--1038}.
\bibitem[{Starzyk and Prasad(2011)}]{starzyk2011computational}
\bibinfo{author}{J.~A. Starzyk}, \bibinfo{author}{D.~K. Prasad},
\newblock \bibinfo{title}{A computational model of machine consciousness},
\newblock \bibinfo{journal}{International Journal of Machine Consciousness}
  \bibinfo{volume}{3} (\bibinfo{year}{2011}) \bibinfo{pages}{255--281}.
\bibitem[{Vassev and Hinchey(2013)}]{vassev2013implementing}
\bibinfo{author}{E.~Vassev}, \bibinfo{author}{M.~Hinchey},
\newblock \bibinfo{title}{Implementing artificial awareness with knowlang},
\newblock in: \bibinfo{booktitle}{Systems Conference (SysCon), 2013 IEEE
  International}, \bibinfo{organization}{IEEE}, pp. \bibinfo{pages}{580--586}.
\bibitem[{Haladjian and Montemayor(2016)}]{haladjian2016artificial}
\bibinfo{author}{H.~H. Haladjian}, \bibinfo{author}{C.~Montemayor},
\newblock \bibinfo{title}{Artificial consciousness and the
  consciousness-attention dissociation},
\newblock \bibinfo{journal}{Consciousness and cognition} \bibinfo{volume}{45}
  (\bibinfo{year}{2016}) \bibinfo{pages}{210--225}.
\bibitem[{Shi et~al.(2017)Shi, Ma, and Li}]{shi2017machine}
\bibinfo{author}{Z.~Shi}, \bibinfo{author}{G.~Ma}, \bibinfo{author}{J.~Li},
\newblock \bibinfo{title}{Machine consciousness of mind model cam},
\newblock in: \bibinfo{booktitle}{International Conference on Knowledge
  Management in Organizations}, \bibinfo{organization}{Springer}, pp.
  \bibinfo{pages}{16--26}.
\bibitem[{Kozma et~al.(2007)Kozma, Aghazarian, Huntsberger, Tunstel, and
  Freeman}]{kozma2007computational}
\bibinfo{author}{R.~Kozma}, \bibinfo{author}{H.~Aghazarian},
  \bibinfo{author}{T.~Huntsberger}, \bibinfo{author}{E.~Tunstel},
  \bibinfo{author}{W.~J. Freeman},
\newblock \bibinfo{title}{Computational aspects of cognition and consciousness
  in intelligent devices},
\newblock \bibinfo{journal}{IEEE Computational Intelligence Magazine}
  \bibinfo{volume}{2} (\bibinfo{year}{2007}) \bibinfo{pages}{53--64}.
\bibitem[{Sun and Franklin(2007)}]{sun2007computational}
\bibinfo{author}{R.~Sun}, \bibinfo{author}{S.~Franklin},
  \bibinfo{title}{Computational models of consciousness: A taxonomy and some
  examples.}, \bibinfo{year}{2007}.
\bibitem[{Gamez(2008)}]{gamez2008progress}
\bibinfo{author}{D.~Gamez},
\newblock \bibinfo{title}{Progress in machine consciousness},
\newblock \bibinfo{journal}{Consciousness and cognition} \bibinfo{volume}{17}
  (\bibinfo{year}{2008}) \bibinfo{pages}{887--910}.
\bibitem[{G{\"o}k and Sayan(2012)}]{gok2012philosophical}
\bibinfo{author}{S.~E. G{\"o}k}, \bibinfo{author}{E.~Sayan},
\newblock \bibinfo{title}{A philosophical assessment of computational models of
  consciousness},
\newblock \bibinfo{journal}{Cognitive Systems Research} \bibinfo{volume}{17}
  (\bibinfo{year}{2012}) \bibinfo{pages}{49--62}.
\bibitem[{Reggia(2013)}]{reggia2013rise}
\bibinfo{author}{J.~A. Reggia},
\newblock \bibinfo{title}{The rise of machine consciousness: Studying
  consciousness with computational models},
\newblock \bibinfo{journal}{Neural Networks} \bibinfo{volume}{44}
  (\bibinfo{year}{2013}) \bibinfo{pages}{112--131}.
\bibitem[{Baars(2002)}]{baars2002conscious}
\bibinfo{author}{B.~J. Baars},
\newblock \bibinfo{title}{The conscious access hypothesis: origins and recent
  evidence},
\newblock \bibinfo{journal}{Trends in cognitive sciences} \bibinfo{volume}{6}
  (\bibinfo{year}{2002}) \bibinfo{pages}{47--52}.
\bibitem[{Sun and Bookman(1994)}]{sun1994computational}
\bibinfo{author}{R.~Sun}, \bibinfo{author}{L.~A. Bookman},
  \bibinfo{title}{Computational architectures integrating neural and symbolic
  processes: A perspective on the state of the art}, volume
  \bibinfo{volume}{292}, \bibinfo{publisher}{Springer Science \& Business
  Media}, \bibinfo{year}{1994}.
\bibitem[{Graziano(2014)}]{graziano2014speculations}
\bibinfo{author}{M.~S. Graziano},
\newblock \bibinfo{title}{Speculations on the evolution of awareness},
\newblock \bibinfo{journal}{Journal of cognitive neuroscience}
  \bibinfo{volume}{26} (\bibinfo{year}{2014}) \bibinfo{pages}{1300--1304}.
\bibitem[{Graziano and Webb(2015)}]{graziano2015attention}
\bibinfo{author}{M.~S. Graziano}, \bibinfo{author}{T.~W. Webb},
\newblock \bibinfo{title}{The attention schema theory: a mechanistic account of
  subjective awareness},
\newblock \bibinfo{journal}{Frontiers in psychology} \bibinfo{volume}{6}
  (\bibinfo{year}{2015}) \bibinfo{pages}{500}.
\bibitem[{Baars(1997)}]{baars1997theatre}
\bibinfo{author}{B.~J. Baars},
\newblock \bibinfo{title}{In the theatre of consciousness. global workspace
  theory, a rigorous scientific theory of consciousness},
\newblock \bibinfo{journal}{Journal of Consciousness Studies}
  \bibinfo{volume}{4} (\bibinfo{year}{1997}) \bibinfo{pages}{292--309}.
\bibitem[{Baars(2005)}]{baars2005global}
\bibinfo{author}{B.~J. Baars},
\newblock \bibinfo{title}{Global workspace theory of consciousness: toward a
  cognitive neuroscience of human experience},
\newblock \bibinfo{journal}{Progress in brain research} \bibinfo{volume}{150}
  (\bibinfo{year}{2005}) \bibinfo{pages}{45--53}.
\bibitem[{Isinkaye et~al.(2015)Isinkaye, Folajimi, and
  Ojokoh}]{isinkaye2015recommendation}
\bibinfo{author}{F.~Isinkaye}, \bibinfo{author}{Y.~Folajimi},
  \bibinfo{author}{B.~Ojokoh},
\newblock \bibinfo{title}{Recommendation systems: Principles, methods and
  evaluation},
\newblock \bibinfo{journal}{Egyptian Informatics Journal} \bibinfo{volume}{16}
  (\bibinfo{year}{2015}) \bibinfo{pages}{261--273}.
\bibitem[{Burke(2002)}]{burke2002hybrid}
\bibinfo{author}{R.~Burke},
\newblock \bibinfo{title}{Hybrid recommender systems: Survey and experiments},
\newblock \bibinfo{journal}{User modeling and user-adapted interaction}
  \bibinfo{volume}{12} (\bibinfo{year}{2002}) \bibinfo{pages}{331--370}.
\bibitem[{Zamberlan et~al.(2017)Zamberlan, Horvath, and Finati}]{TCCMiller}
\bibinfo{author}{M.~Zamberlan}, \bibinfo{author}{M.~Horvath},
  \bibinfo{author}{V.~Finati},
\newblock \bibinfo{title}{Sistema preditivo de recomendacao baseado em filtros
  colaborativos e clusterizacao},
\newblock \bibinfo{address}{São Bernardo do Campo}, \bibinfo{year}{2017}.
\bibitem[{Horvath et~al.(2018)Horvath, Zamberlan, Finati, Wachs-lopes, and
  Rodrigues}]{2018-zamberlan}
\bibinfo{author}{M.~Horvath}, \bibinfo{author}{M.~Zamberlan},
  \bibinfo{author}{V.~Finati}, \bibinfo{author}{G.~A. Wachs-lopes},
  \bibinfo{author}{P.~S. Rodrigues},
\newblock \bibinfo{title}{{Exploiting Cluster Specialization into Linear
  Weighted Hybrid Recommender Systems}}  (\bibinfo{year}{2018}).

\end{thebibliography}








\end{document}